\documentclass[11pt,a4paper]{article}
\usepackage{times,latexsym}
\usepackage{url}
\usepackage[english]{babel}
\usepackage{graphicx}
\usepackage{amsmath}
\usepackage{amsthm}
\usepackage{amssymb}
\usepackage{url}
\usepackage{array}
\usepackage{textcomp}
\usepackage{listings}
\usepackage{hyperref}
\usepackage{xcolor}
\usepackage{colortbl}
\usepackage{float}
\usepackage{gensymb}
\usepackage{caption}

\usepackage[T1]{fontenc}

\usepackage[]{tacl2018v2}

\taclpubformattrue

\usepackage{xspace,mfirstuc,tabulary}

\newif\iftaclinstructions
\taclinstructionsfalse 
\iftaclinstructions
\renewcommand{\confidential}{}
\renewcommand{\anonsubtext}{(No author info supplied here, for consistency with
TACL-submission anonymization requirements)}
\newcommand{\instr}
\fi

\iftaclpubformat 

\else

\fi

\title{Using Machine Learning and Natural Language Processing Techniques to Analyze and Support Moderation of Student Book Discussions}

\author{
 Jernej Vivod \\
 Faculty of Computer and Information Science, University of Ljubljana \\
 Ljubljana, 1000, Slovenia \\
  {\sf vivod.jernej@gmail.com} \\
}

\date{\today}

\begin{document}
\maketitle
\begin{abstract}
The increasing adoption of technology to augment or even replace traditional face-to-face learning has led to the development of a myriad of tools and platforms aimed at engaging the students and facilitating
the teacher's ability to present new information. The IMapBook project aims at improving the literacy and 
reading comprehension skills of elementary school-aged children by presenting them with interactive 
e-books and letting them take part in moderated book discussions. This study aims to develop and 
illustrate a machine learning-based approach to message classification that could be used to 
automatically notify the discussion moderator of a possible need for an intervention and also to collect
other useful information about the ongoing discussion. We aim to predict whether a message posted in the
discussion is relevant to the discussed book, whether the message is a statement, a question, or an answer,
and in which broad category it can be classified. We incrementally enrich our used feature subsets and
compare them using standard classification algorithms as well as the novel Feature stacking method. 
We use standard classification performance metrics as well as the Bayesian correlated t-test to show 
that the use of described methods in discussion moderation is feasible. Moving forward, we seek to 
attain better performance by focusing on extracting more of the significant information found in the 
strong temporal interdependence of the messages.
\end{abstract}

\section{Introduction}

Recent decades have brought about an increase in the use of computer-based tools in practically every 
field of human endeavor. The field of education is no exception. Such tools can be used to augment or 
even completely replace traditional face-to-face teaching methods. The emergence of online learning
platforms has necessitated the development of means to enable learning activities, such as 
group discussions, to be performed through the use of technology. One such example of a learning 
platform is the IMapBook software suite aimed at increasing the literacy and reading 
comprehension skills of elementary school-aged children through the use of web-based eBooks, 
embedded games related to their contents, as well as moderated group discussions.
Keeping these discussions constructive and relevant can be difficult and usually requires a 
discussion moderator to be present at all times. This
can limit the opportunities for such discussions to take place. Leveraging the methods and insights 
from the fields of artificial intelligence and
machine learning, we can attempt to develop systems to automatically classify messages into 
different categories and detect when the discussion has
veered off course and necessitates intervention. Our research tackles this problem using a 
compilation of discussions obtained during pilot studies
testing the effectiveness of using the IMapBook software suite in 4th-grade classrooms. 
The studies were performed in 8 different Slovene primary schools and, in total, included 342 students. 
The discussions consist of 3541 messages along with annotations specifying their relevance to the 
book discussion, type, category, and broad category. The ID of the book being discussed and the time 
of posting are also included, as are the poster's school, cohort, user ID, and username. 
Each message was also manually translated into English to aid non-Slovene-speaking researchers. 
The use of the Slovene language presents unique challenges in applying standard language 
processing methods, many of which are not as readily available as for other, more widely spoken languages.

Given a sequence of one or more newly observed messages, we want to estimate the relevance of 
each message to the actual topic of discussion. Namely,
we want to assign messages into two categories — relevant to the book being discussed or not. 
Additionally, we want to predict whether the message is
a question, an answer, or a statement which we call the type of the message. Finally, we want to 
assign a category label to each message where the
possible labels can be either 'chatting', 'switching', 'discussion', 'moderating', or 'identity'. 
Building a predictive model capable of performing
such predictions with acceptable performance would allow us to experiment with including this new 
level of automation in the IMapBook software suite
as well as in any related products. The research insights are also applicable to areas such as 
online user comments and content moderation.

\section{Related Work}

The objective of our research is closely related to tasks concerning online 
content moderation which has been the subject of
much research in recent years. Perhaps one of the earliest studies done on this subject is the 2009 
study by Yin et al. \cite{Yin2009DetectionOH} in which the authors used sentiment/contextual features 
in tandem with the TF-IDF approach to detect online harassment. An earlier study by Mclaren et al. specifically focuses on the use of machine learning techniques to support the mediation of student 
online discussions in a uniquely constrained network-like environment that differs significantly from 
ours \cite{mediation}. A 2016 study by Kadunc focuses on using machine learning methods to analyze 
the sentiment of Slovene online comments and provides an important contribution in the form of an 
opinion lexicon \cite{KADUNC_2016}. However, the specifics and unique challenges presented by the 
problem of classifying short Slovene text produced by this age group remains an area with little to 
no research currently done.

\section{Methods}

Achieving the goal of creating a working predictive model for the task of message classification 
requires careful processing of the raw data in such a way as to expose as much useful information 
as possible. This process of feature extraction and feature engineering often results in very 
high dimensional descriptions of our data that can be prone to problems arising as part of the 
so-called curse of dimensionality \cite{Domingos_afew}. This can be mitigated by using 
classification models well-suited for such data as well as performing feature 
ranking and feature selection.

\subsection{Extracting and Engineering Features}

Building a quality predictive model requires a good characterization of each message in 
terms of discriminative and non-redundant features. Extracting such features from raw 
text data is a non-trivial task that is subject to much research in the field of 
natural language processing. Here, we describe the feature extraction process used in our study. 
A detailed evaluation of the features is presented in the Results section.

\subsubsection{General Message Features}

Looking at the messages in our dataset, we can immediately notice simple but potentially important
differences between the messages in terms of word count, punctuation use, and other attributes that can
be easily deduced by merely inspecting the raw data without any need for context.

We extract the number of words in the message, the maximal, minimal, and 
average word lengths, the number of digits in the message,
the number of punctuation marks in the text, the number of capital letters in the text, 
the number of consecutive 
repeated characters and check whether the message starts with a capital letter 
and if it ends with a period.

\subsubsection{Important Words}

We can gain valuable insight by observing the presence of members from important 
word groups. We compiled lists of chat usernames
used in the discussions, common given names in Slovenia, common curse words used in 
Slovenia as well as any proper names found
in the discussed books. We also created a list of nouns, verbs and adjectives which we 
observed to be highly discriminative such 
as \textit{misliti} (to think), \textit{knjiga} (book), \textit{najljubše} (favorite) 
among others. We noted any presence or absence of words from these lists as features 
describing the pertaining message.

\subsubsection{Multiset-Based Features}

The Bag-of-words model and its variations characterize documents by counting the occurrences of 
each word from a pre-defined set.
It is based on the simple assumption that similar documents share a similar vocabulary. 
We can augment the basic Bag-of-words
model by weighing each word in the vocabulary proportional to its rarity in the corpus, 
using the assumption that rarity implies discriminativity.

We constructed a simple bag-of-words model using unigrams and bigrams. We required both
the unigrams and bigrams to appear in at least two messages.
All words were converted to lower case, any punctuation and emojis were disregarded and the words
were lemmatized before the construction of the model. We also converted any consecutive repeated
letters in the words to single occurrences. 
The resulting model consisted of 2009 unique unigrams and bigrams. We used a similar method
to include part-of-speech tagging into our set of considered features. 
We describe this process in detail in the following subsection.

\subsubsection{Part-of-Speech Tagging-Based Features}

Part-of-speech tagging is the process of labeling the words in a text based on their 
corresponding part of speech. Describing each message in terms of its associated part-of-speech 
labels allows us to use another perspective from which we can view
and analyze the corpus. The use of non-standard Slovene and misspellings make part-of-speech 
tagging a non-trivial task. We
attempted to solve the issue by constructing a simple dictionary mapping known non-standard 
and colloquial versions of Slovene words into their standard equivalents. 
We constructed the dictionary using the corpus available as part of
the JANES project \cite{janes}. Again, we removed any punctuation, 
repeated characters, and converted the text to lower-case before
applying the dictionary. We used a part-of-speech tagger trained on the IJS JOS-1M corpus to perform
the tagging \cite{tagger}. We simplified the results by considering only the part of 
speech and its type. We characterized each message by the number of occurrences of each label 
which can be viewed as applying a bag-of-words model with 'words' being the part-of-speech tags.

\subsubsection{Time-Based Features and Models}

The features described so far consider each message as an independent unit and do not take 
into consideration the inherent time dependency. By observing the temporal distribution 
of labels, we can observe that the labels are not uniformly distributed. Messages
relevant to the book seem to appear in clusters and observing a message marked as a 
question naturally leads us to expect an answer in the subsequent messages. We can use the sequence 
of labels in the dataset to compute a label transition probability matrix defining a Markov model. 
We can also compute the conditional probabilities of each label based on the labels of the
previous messages. We implemented both models with conditional probabilities computed given 
the previous 4 labels. We also assigned to each message the number of times the poster has posted 
in a row and the number of messages authored by the posters in the last 20 messages.

\subsection{Class Label Distribution and Resampling}

It is important to inspect the distribution of class labels in any dataset and note any severe 
imbalances that can cause problems in the model construction phase as there may not be enough data 
to accurately represent the general nature of the underrepresented group. Such imbalances also 
warrant care in result interpretation.

\begin{figure}[!htb]
    \centering
    \includegraphics[width=1.0\columnwidth]{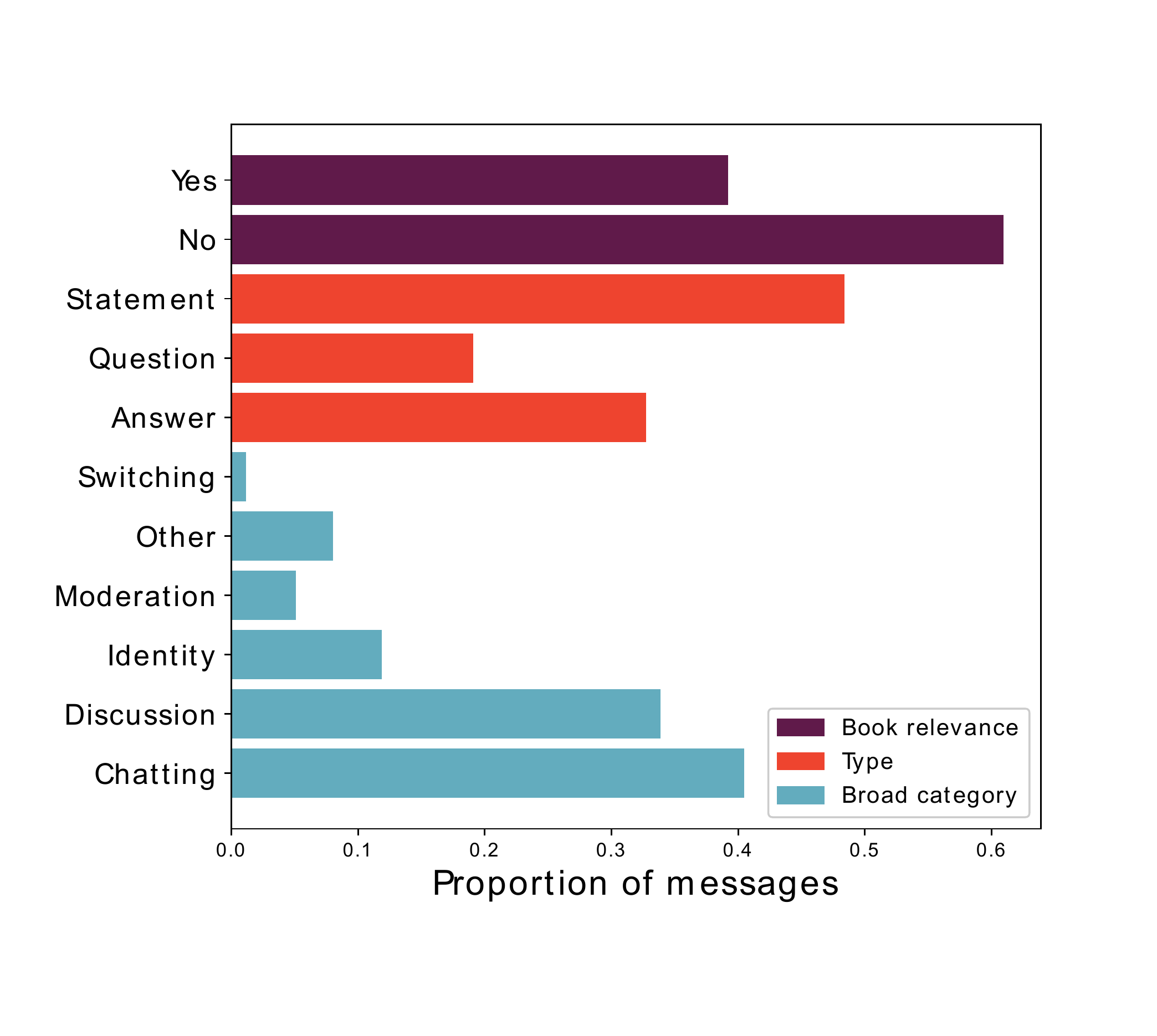}
    \caption{Class label distribution for different prediction objectives.}
    \label{fig:label_dist}
\end{figure}

Figure \ref{fig:label_dist} shows the distribution of class labels for each of the prediction objectives.
We can see that the distribution of broad category labels is notably imbalanced with 40.3\% of messages
assigned to the broad category of 'chatting', but only 1\%, 4.5\% and 8\% to 'switching', 'moderation'
and 'other' respectively.

Several methods have been developed to handle such imbalances, the most notable being resampling and
the use of cost-sensitive classification methods. Resampling techniques aim to balance 
the dataset by removing data from the overrepresented classes, generating synthetic data for the minority
classes, or both. We evaluate the use of SMOTE resampling in tandem with cleaning 
unwanted overlap between classes using TOMEK links \cite{tomek}.

\subsection{Feature Ranking and Feature Selection}

The relative values of features for building a quality predictive model 
often vary significantly. Determining the importance
of features is a non-trivial task for which many metrics and methods have 
been proposed. The subset of features used to build
the model can have an important effect on its performance and overall usefulness. 
A model induced on a well-chosen feature subset
will be more general and easier to interpret. A notable group of algorithms well-suited 
for the task of ranking features is
the family of Relief-like algorithms which offer performance acceptable for use with large 
datasets as well as the notable
capability of detecting feature-feature interactions. These algorithms work by sampling training 
data instances and scoring
the attributes based on how well they separate the sampled instances from closest instances 
corresponding to a different class
as well as on the similarity to closest instances from the same class by this 
attribute \cite{Kononenko1997}. We use the SWRF* (Sigmoid Weighted
ReliefF Star) algorithm to perform the feature ranking \cite{swrf}. 
We also estimate the importance of features by observing the coefficients of a 
fitted logistic regression model and analyzing the Gradient boosting model fitted to the training data.

\subsection{Prediction Models}

The rapidly advancing field of machine learning has produced a myriad of methods that can be 
used to make predictions in a supervised learning setting. We evaluate the use of well-known 
classification algorithms such as Random forests, Support vector machines, Gradient boosting, 
and logistic regression. We also
implement and evaluate an ensemble method called Feature stacking that was specially 
developed for such tasks. It is important to critically compare any results obtained by 
such sophisticated methods to the outputs of baseline models such as the 
so-called majority classifier, which always predicts the most common label found in training
data with maximal certainty as well as the random-guess or uniform classifier which, 
predicts the labels uniformly at random. To be useful, any implemented method should be 
statistically proven to outperform these trivial baselines.

\subsubsection{The Feature Stacking Method}

A notable algorithm developed for the task of sentence classification that is especially 
suited for high-dimensional data is the so-called Feature stacking approach \cite{lui-2012-feature} 
which combines several classifiers each trained on a feature subset and a final classifier that 
takes as its input the output of these classifiers. The authors of the paper describing the 
Feature stacking approach suggest using logistic regression as the feature classification method 
since it makes the method resemble a neural network structure. We use an SVM as the final classifier.

\begin{figure}[!htb]
    \centering
    \includegraphics[width=1.0\columnwidth]{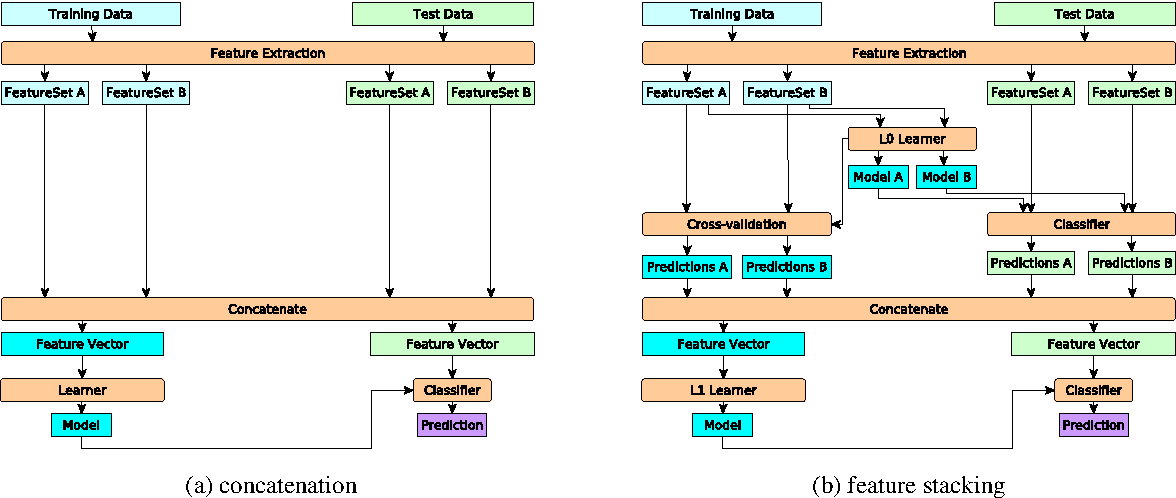}
    \caption{The feature stacking method contrasted with the typical method of concatenating features (taken from \cite{lui-2012-feature}).}
    \label{fig:feat_stacking}
\end{figure}

During training, the training data is converted to new features consisting of logistic regression 
outputs for each feature subset. This is achieved using k-fold cross-validation. 
Next, logistic regression is fitted to the entire training data feature subsets and is used to 
encode the test data. A final meta-classifier is fitted to the training data encoded using 
logistic regression. Test data is first encoded using a trained logistic regression model and 
finally classified with the meta-classifier. The features stacking method is contrasted with the 
more typical feature concatenation method on the diagram shown in figure \ref{fig:feat_stacking}.

\subsubsection{Utilizing Predictions of Temporal Models}

We combined the predictions of the classification model with the probabilities computed using the Markov model and the
conditional probabilities by weighing the class probabilities obtained by each method as shown below.
\begin{align*}
    p(C) = &(1 - \alpha - \beta) \cdot p_{c}(C) \\ &+ \alpha \cdot p_{m}(C) \\
    &+ \beta \cdot p(C|\textrm{previous n labels})
\end{align*}
Here, $p_{c}$ and $p_{m}$ represents the probability obtained by the classification model and the Markov model respectively.
We performed an exhaustive cross-validated grid-search to tune the $\alpha$ and $\beta$ parameters of the combined ensemble model.

\section{Experiments}

\subsection{Experimental Setup}

We began by building classification models using only general features obtained by observing the 
word counts, word lengths, and character properties in individual messages. 
We compared the use of different models and performed
feature scoring to rank the perceived usefulness of each feature. All model 
evaluations were performed using 10 repetitions of 10-fold cross-validation. 
Explicit comparisons between different methods were made using the Bayesian correlated
t-test which can be used to compute probabilities of one method being better than the other and 
avoids some common pitfalls associated with the more typical frequentist
approaches~\cite{benavoli2016time}. Subsequently, the next batch of features was added to the feature
extraction/engineering process and the evaluation process repeated. for the initial feature subset
evaluations, we focused exclusively on the book relevance prediction objective. 
Using the full feature set, we evaluate the best scoring models on all prediction objectives.

\subsection{Experimental Results}

\subsubsection{The Initial Feature Subset}

Table \ref{Tab:res1} shows the results obtained by evaluating the support vector machine model
built using the starting set of features. Using the Bayesian correlated
t-test, we estimated the support vector machine model to be better than the random forest model with 
a probability equal to $0.59$. The probability of the models performing equally was estimated to be $0.39$.
The support vector machine was also estimated to be marginally better than the
gradient boosting model with a probability of 0.48 and a probability of 0.47 of them being equal. 
We did not use the feature stacking method as it is not well defined for data with few features.
The model outperformed the baseline random and majority models with an estimated probability of 1.0.

\begin{table}[!htb]
\begin{tabular}{l|l|l}
          & not relevant & relevant \\ \hline
precision & 0.801         & 0.696     \\
recall    & 0.807         & 0.691     \\
f1-score  & 0.798         & 0.702     \\
support   & 215.5        & 138.6   
\end{tabular}
\caption{Results for the book relevance classification objective obtained by 
evaluating the support vector machine model using the initial feature subset.}
\label{Tab:res1}
\end{table}

We used the gradient boosting algorithm, logistic regression coefficients, and the SWRF* 
algorithm to estimate the discriminativity of features in the initial subset. 
Averaging the estimations, the average word length and the word count of the message 
were deemed most important, followed by the maximal word length and the amount of 
punctuation in the message.

\begin{figure}[!htb]
    \centering
    \includegraphics[width=1.0\columnwidth]{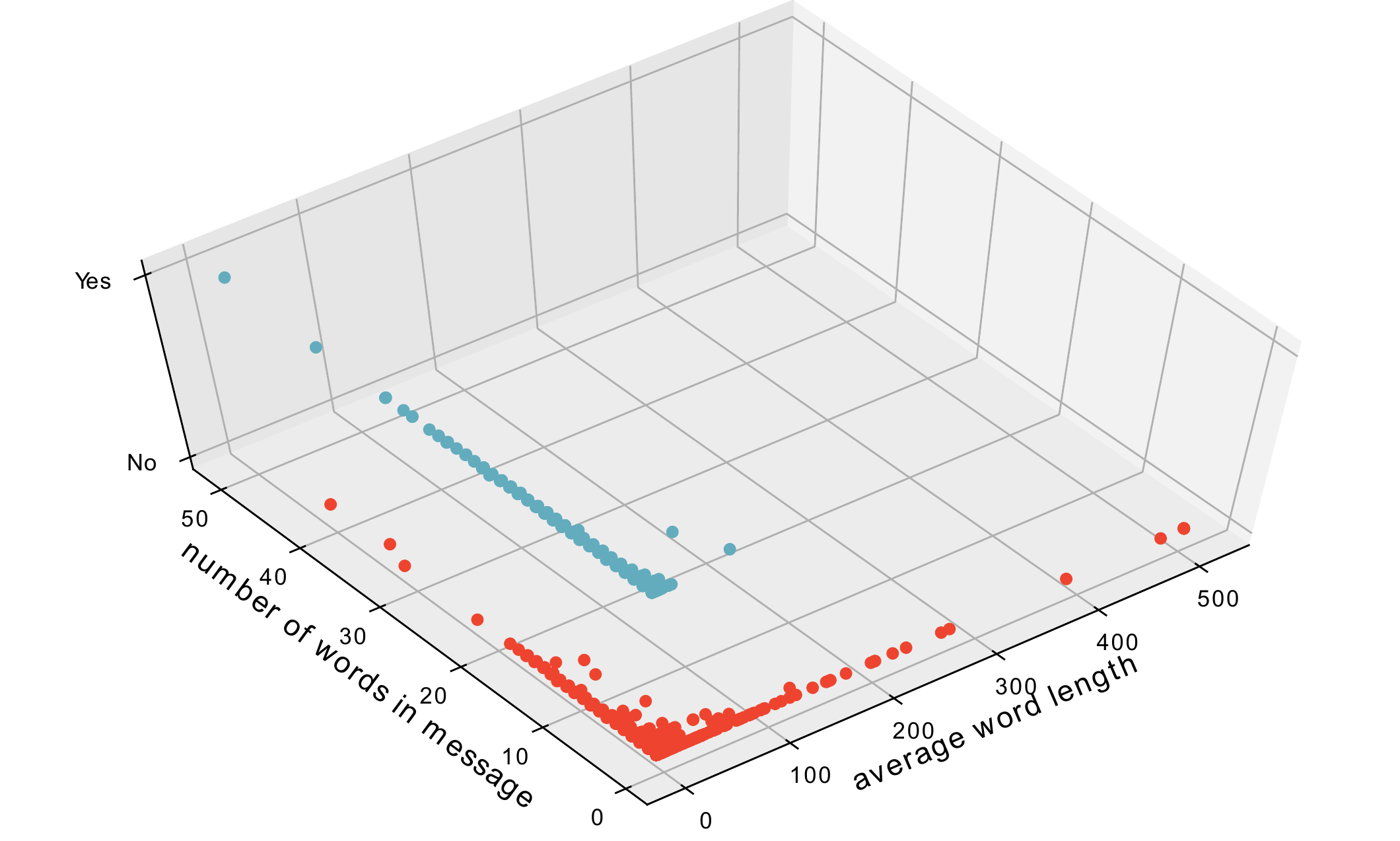}
    \caption{Seperability of relevant and non-relevant messages by average word length and word count.}
    \label{fig:sep1}
\end{figure}

Figure \ref{fig:sep1} shows the separability of relevant and non-relevant messages by the two 
top-rated features. We can see that a notable portion of non-relevant messages has a 
considerably higher average word length. This can be explained by observing that many 
non-relevant messages contain long gibberish words as well as words with consecutively 
repeated letters.

\subsubsection{Including Important Words and User's Post History}

We augmented the initial feature subset with counts of curse words, 
repeated letters, counts of special verbs and nouns deemed important, 
such as 'misliti' (to think), 'knjiga' (book), counts of common Slovene given names, 
counts of chat usernames, the number of times the poster posted in a row and the portion of poster's
posts in the last 20 messages. We repeated the evaluation process for the book 
relevance prediction objective. Using the Bayesian correlated t-test, the SVM classifier was 
again evaluated to be better by a small margin.

Table \ref{Tab:res2} shows the results obtained by evaluating the support vector 
machine model build using the augmented set of features. 

\begin{table}[!htb]
\begin{tabular}{l|l|l}
          & not relevant & relevant \\ \hline
precision & 0.825         & 0.731     \\
recall    & 0.828         & 0.727     \\
f1-score  & 0.826         & 0.728     \\
support   & 215.5        & 138.6   

\end{tabular}
\caption{Results for the book relevance classification objective obtained by 
evaluating the support vector machine model using the augmented feature subset.}
\label{Tab:res2}
\end{table}

Using the Bayesian correlated t-test, the support vector machine model built using
the augmented feature set was estimated to be better than the one built using the
initial feature set with a probability of 0.97. The probability of them being equal 
was estimated to be 0.02.

\subsubsection{Including Bag-of-Words Features}

We proceeded by adding the unigram and bigram counts to the feature set. 
This increased the dimensionality of the dataset substantially by adding 2009 additional 
features. Using the Bayesian correlated t-test, the feature stacking method was 
determined as the most probable best classification model.

Table \ref{Tab:res3} shows the results obtained by evaluating the feature stacking method model 
build using the enriched set of features. 

\begin{table}[!htb]
\begin{tabular}{l|l|l}
          & not relevant & relevant \\ \hline
precision & 0.872         & 0.851     \\
recall    & 0.918         & 0.706     \\
f1-score  & 0.894         & 0.772      \\
support   & 215.5        & 138.6   

\end{tabular}
\caption{Results for the book relevance classification objective obtained by 
evaluating the feature stacking method model using additional bag-of-words features.}
\label{Tab:res3}
\end{table}

The estimated probability that the new feature set produces a better model is equal to 1.0.

\subsubsection{Including Part-Of-Speech Tagging Features}

Next, we included the Part-of-Speech tagging based features consisting of 
the part of speech and its type pair counts. This added 30 additional features. Again,
the most probable best model was evaluated to be the feature stacking method.

The comparison between feature stacking method models either using POS tagging-based features or not indicates that the new features do not improve the
model for this prediction objective. The probability of the model built using the previous feature set being better is estimated to be 0.25 while the probability of the models being equal is estimated to be 0.67.

\subsubsection{Including the Temporal Models}

\begin{figure}[!htb]
    \centering
    \includegraphics[width=1.0\columnwidth]{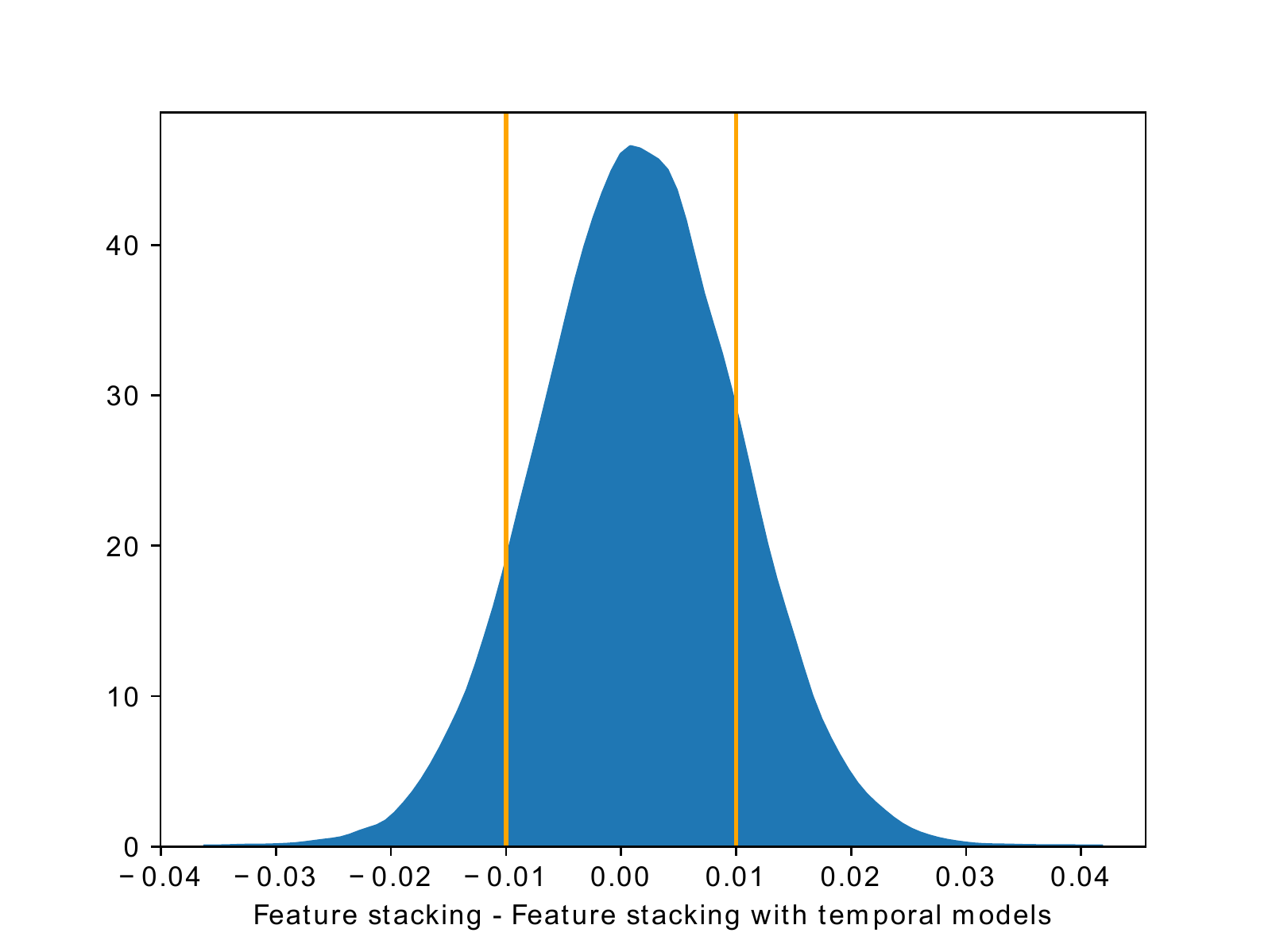}
    \caption{Posterior distribution used to compute the probabilities of one 
    classification model being better than the other. The region of practical equivalence 
    is denoted by the vertical bars.}
    \label{fig:bctt1}
\end{figure}

We used an exhaustive, cross-validated grid search to determine the optimal weights of the 
Markov chain model and the conditional probability-based model, which were estimated to be 0.06 
and 0.07 respectively. We compared the feature stacking model with and without the combined use of 
temporal models using the Bayesian correlated t-test for the book relevance prediction objective. 
The ensemble model combining the predictions of temporal models was evaluated to be better with a
probability of 0.16 while the probability of the models being equal was estimated to be 0.74. 
The posterior distribution used to compute the probabilities is shown in figure \ref{fig:bctt1}.

\subsubsection{Evaluating the Models using All Prediction Objectives}

\begin{figure}[!htb]
    \centering
    \includegraphics[width=0.7\columnwidth]{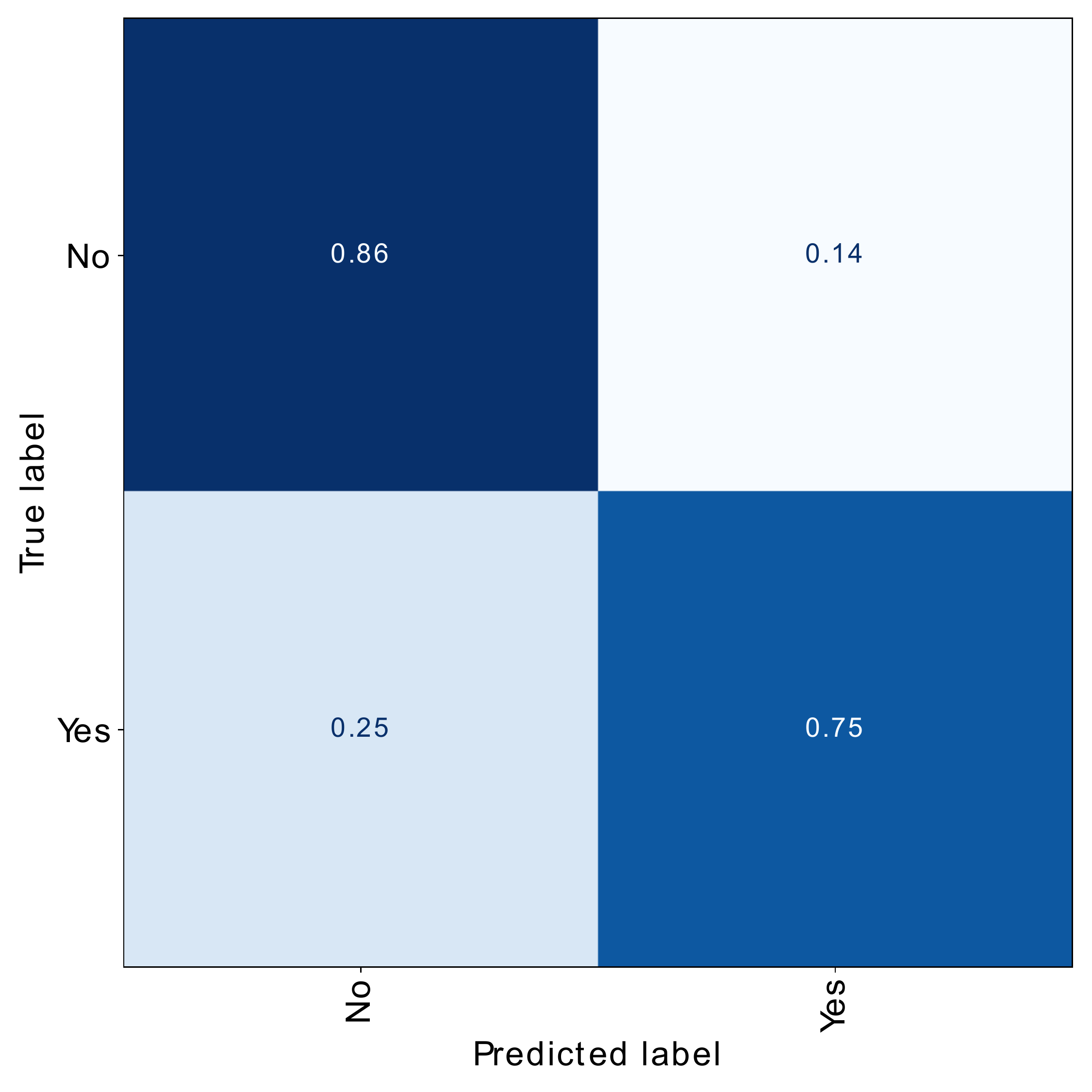}
    \caption{Confusion matrix obtained using a 80/20 train-test split.}
    \label{fig:cm1}
\end{figure}

Figure \ref{fig:cm1} shows the confusion matrix for the book relevance prediction objective using 
an 80/20 train-test split and the feature stacking method model. It is useful to inspect the 
messages corresponding to false negatives or false positives and compare them to correctly 
classified messages to try to determine ways the feature set could be improved. Table \ref{Tab:fnfp} 
lists a subset of false negatives and false positives. We can see that the actual label can be 
extremely dependent on the context of the conversation which makes it very difficult for a model 
with limited ability to process such context to correctly classify messages shown in the table.

\begin{table}[!htb]
\tiny
\begin{tabular}{l|l|l}
Message                                                & Predicted & True \\ \hline
zakaj                                                  & no              & yes        \\
aja                                                    & no              & yes        \\
prestrašeno                                            & no              & yes        \\
itak                                                   & no              & yes        \\
MMMMMMMMMMMMMMMMM NE VEM                               & no              & yes        \\
Jakob, kaj pomeni razdal? Nadaljuj z branjem, prosim.  & yes             & no         \\
AMPAK POTEM SI UMAZAN                                  & yes             & no         \\
Odlična ideja, Smartno B19, prosim nadaljuj z branjem. & yes             & no         \\
SmartnoB23, kaj pa ti meniš o odgovoru na vprašanje?   & yes             & no         \\
jaz se tudi strinjam z smartno 11                      & yes             & no        
\end{tabular}
\caption{A subset of false negatives and false positives for the book relevance prediction objective.}
\label{Tab:fnfp}
\end{table}

\begin{figure}[!htb]
    \centering
    \includegraphics[width=1.0\columnwidth]{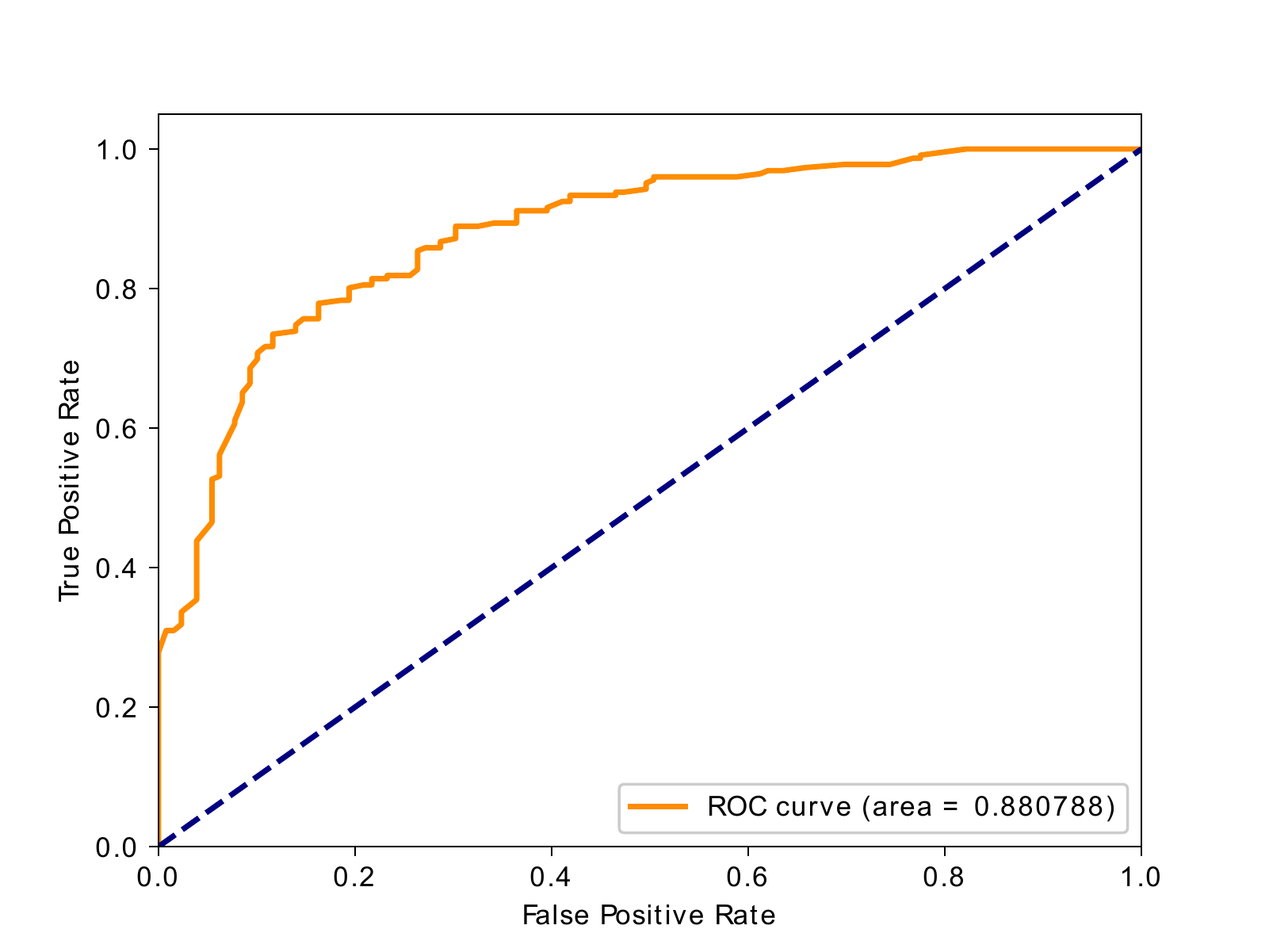}
    \caption{ROC curve and the AUROC value obtained using a 80/20 train-test split.}
    \label{fig:roc}
\end{figure}

Figure \ref{fig:roc} shows the ROC curve and the AUROC value obtained using the same 
train-test split and model.

We proceeded by evaluating the models using the type and broad category 
prediction objectives using the full feature set. We report the results
for the feature stacking method which was estimated by the Bayesian correlated t-test to have 
the highest probability of being the best model in the evaluated set of models.

Table \ref{Tab:res4} shows the results obtained by evaluating the feature stacking method for 
the message type prediction objective.

\begin{table}[!htb]
\begin{tabular}{l|l|l|l}
          & A & Q & S\\ \hline
precision & 0.708 & 0.773 & 0.784    \\
recall    & 0.734 & 0.706 & 0.788    \\
f1-score  & 0.718 & 0.737 & 0.785    \\
support   & 115.7 & 67.3  & 171.1 
\end{tabular}
\caption{Results for the type classification objective obtained by evaluating the 
feature stacking method model using the complete feature set.}
\label{Tab:res4}
\end{table}

\begin{figure}[!htb]
    \centering
    \includegraphics[width=0.68\columnwidth]{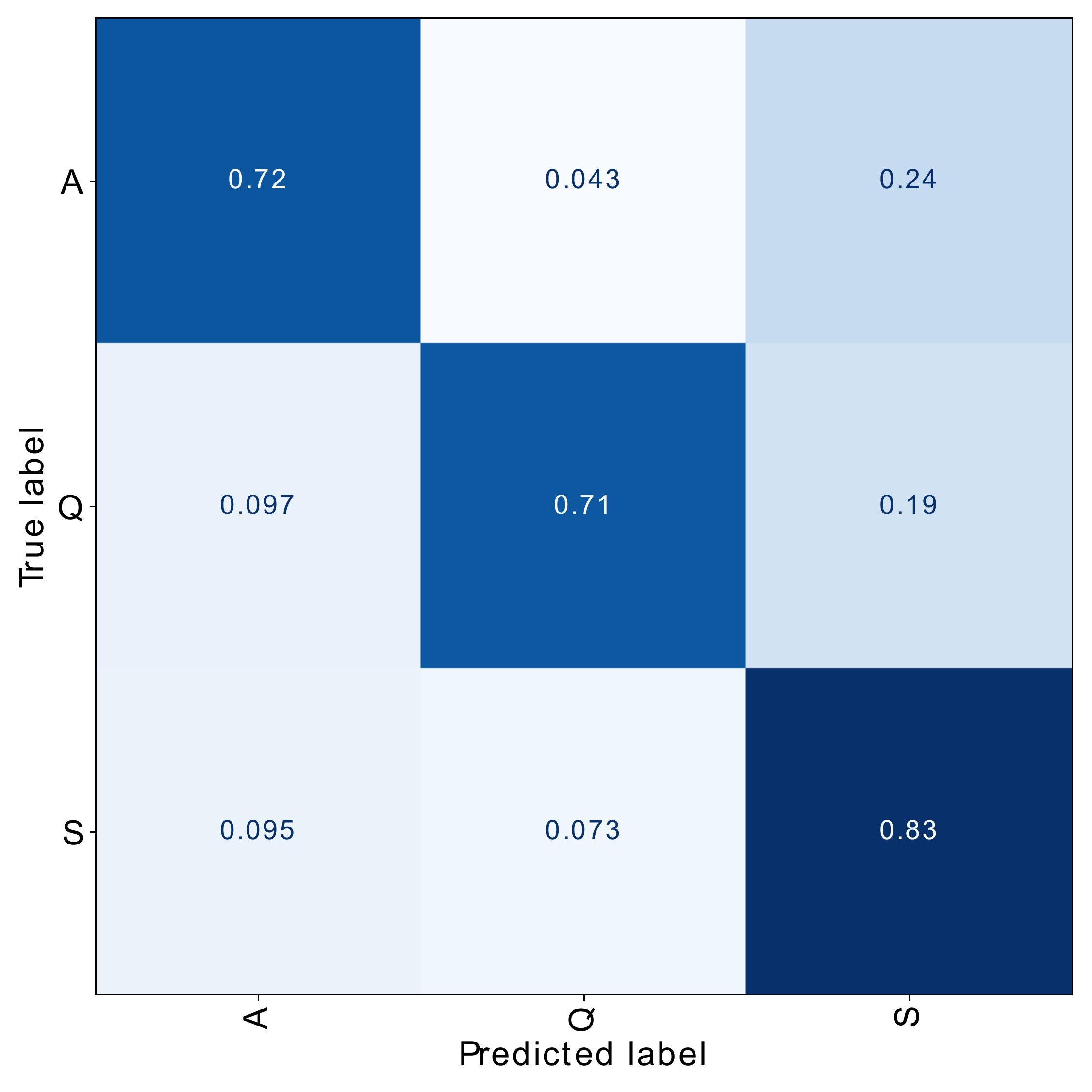}
    \caption{Confusion matrix obtained using a 80/20 train-test split. 'S' refers to the 'statement' label, 'Q' to the 'question' label, and 'A' to the 'answer' label.}
    \label{fig:cm2}
\end{figure}

Figure \ref{fig:cm2} shows the confusion matrix obtained using a train-test split 
for this classification objective.

Table \ref{Tab:res5} shows the results obtained by evaluating the feature stacking method 
for the broad category prediction objective. 

\begin{table}[!htb]
\scriptsize
\begin{tabular}{l|l|l|l|l|l|l}
          & C & D & I & M & O & S\\ \hline
precision & 0.632 & 0.683 & 0.895 & 0.878 & 0.859 & 0.0 \\
recall    & 0.766 & 0.796 & 0.496 & 0.360 & 0.321 & 0.0 \\
f1-score  & 0.691 & 0.735 & 0.635 & 0.503 & 0.460 & 0.0 \\
support   & 143.0 & 119.7  & 41.7 & 17.7  & 28.2  & 3.8
\end{tabular}
\caption{Results for the broad category classification objective obtained by evaluating the feature stacking method model using the complete feature set.}
\label{Tab:res5}
\end{table}

\begin{figure}[!htb]
    \centering
    \includegraphics[width=0.7\columnwidth]{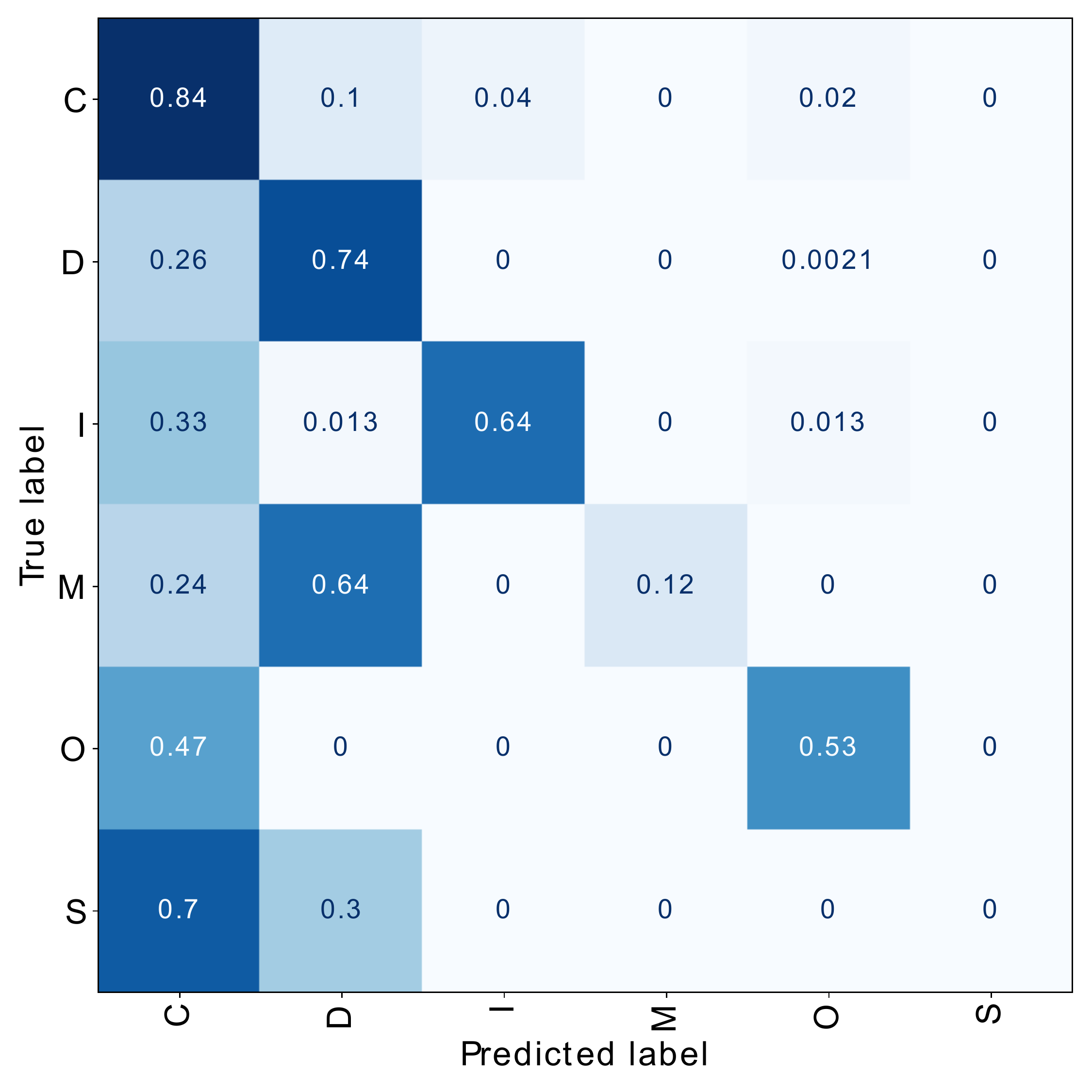}
    \caption{Confusion matrix obtained using a 80/20 train-test split. 'C' refers to
    the 'chatting' label, 'D' refers to the 'discussion' label, 'I' refers to the 'identity' label, 'M' refers to the 'moderation' label, 'O' refers to the 'other' label, and 'S' refers to the 'switching' label.}
    \label{fig:cm3}
\end{figure}

Figure \ref{fig:cm3} shows the confusion matrix obtained using a train-test split for 
this classification objective. As stated earlier, it should be noted that the label distribution for 
this prediction objective is notably imbalanced. Consequently the proportion of samples 
labeled 'moderation', 'other', and 'switching' in the test set was 6.3\%, 4.7\%, and 3.2\% respectively.

All models outperformed the baselines with the estimated probability of 1.0.

\section{Conclusion and Future Work}

The best results were achieved by using the Feature stacking method model built on the complete 
feature subset. The results indicate the performance to be sufficient for the methods to be used 
in real-world tools and platforms. A significant portion of the information needed for 
correct classifications is hidden in the strong temporal interdependence of the messages which 
our developed methods exploited only marginally.

\bibliography{literature}
\bibliographystyle{acl_natbib}

\end{document}